\documentclass[11pt,a4paper]{article}
\usepackage[hyperref]{naaclhlt2019}
\usepackage{times}
\usepackage{latexsym}

\usepackage{url}
\usepackage{graphicx}
\usepackage{xcolor}
\usepackage{colortbl}
\definecolor{aliceblue}{rgb}{0.94, 0.97, 1.0}
\definecolor{airforceblue}{rgb}{0.36, 0.54, 0.66}
\definecolor{grannysmithapple}{rgb}{0.66, 0.89, 0.63}
\newcommand{\HeaderColor}{\rowcolor{grannysmithapple!60}}
\newcommand{\RowColor}{\rowcolor{aliceblue!75}}

\usepackage{subcaption}
\usepackage{amsmath}
\mathchardef\mhyphen="2D 

\aclfinalcopy 
\def\aclpaperid{8} 

\title{Referring to Objects in Videos using\\ Spatio-Temporal Identifying Descriptions}

\author{\:\:\: Peratham Wiriyathammabhum\textsuperscript{$\spadesuit$$\diamondsuit$}, Abhinav Shrivastava\textsuperscript{$\spadesuit$$\diamondsuit$}, \textbf{Vlad I. Morariu\textsuperscript{$\diamondsuit$}}, \textbf{Larry S. Davis\textsuperscript{$\spadesuit$$\diamondsuit$}} \\ 
 University of Maryland: Department of Computer Science$\spadesuit$, UMIACS$\diamondsuit$ \\
  {\tt peratham@cs.umd.edu, abhinav@cs.umd.edu}\\ {\tt morariu@umd.edu, lsd@umiacs.umd.edu} \\
  }

\begin{document}
\maketitle
\begin{abstract}
  This paper presents a new task, the grounding of spatio-temporal identifying descriptions in videos. Previous work suggests potential bias in existing datasets and emphasizes the need for a new data creation schema to better model linguistic structure. We introduce a new data collection scheme based on grammatical constraints for surface realization to enable us to investigate the problem of grounding spatio-temporal identifying descriptions in videos. We then propose a two-stream modular attention network that learns and grounds spatio-temporal identifying descriptions based on appearance and motion. We show that motion modules help to ground motion-related words and also help to learn in appearance modules because modular neural networks resolve task interference between modules. Finally, we propose a future challenge and a need for a robust system arising from replacing ground truth visual annotations with automatic video object detector and temporal event localization. 
\end{abstract}
\section{Introduction}
Localizing referring expressions in videos involves \textit{both static and dynamic information}. A referring expression \cite{dale1995computational, roy2005connecting} is a linguistic expression that grounds its meaning to a specific referent object in the world. The input video can be very long, have unknown length, contain many objects from the same class, or contain similar actions and interactions throughout the video. A successful, grounded communication between a speaker and a listener must ensure that the sentence or discourse provides enough information such that the listener can eliminate all distractors and focus only on the referent object that acts in a specific time interval. That essential information varies from the diversity of events in the world. However, a speaker is likely to mention salient properties and also salient differences based on the referent in comparison to other distractors. The differences can be about object category, attributes, poses, actions, changes in location, relationships and contexts in the scene.

\begin{figure}[t] \label{fig1}
\begin{center}
   \includegraphics[width=0.90\linewidth]{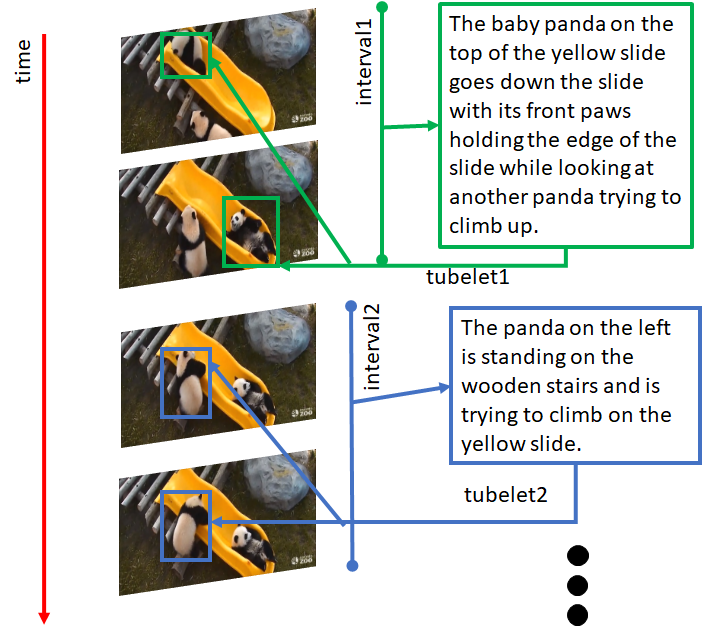}
\end{center}
   \caption{The first spatio-temporal identifying description in the green box grounds to the event that a panda goes down the slide. Another panda can be a context because they are interacting in the same scene. The second identifying description in the blue box grounds to the event that another panda climbs up the slide.}
\label{fig:long}
\label{fig:onecol}
\vspace{-1\baselineskip}
\end{figure}

Existing \textit{image referring expression} datasets  \cite{mao2016generation, johnson2015image, kazemzadeh2014referitgame, plummer2017flickr30k, krishna2017visual} do not contain referring expressions that refer to dynamic properties or movements of the referent. These datasets do not require temporal understanding that would require a system to learn that ``moving to the right'' is different from ``moving to the left'' and ``getting up'' is different from ``lying down''. Existing \textit{video referring expression} datasets and approaches \cite{krishna2017dense, hendricks17iccv, gao2017tall, berzak2015you, li2017tracking, hendricks2018localizing, gavrilyuk2018actor} focus only on temporal localization but referent object localization. In other words, they do not ground events in both space and time. Emphasized by \cite{cirik2018visual}, the data collection process for referring expressions should incorporate linguistic structure such that the model can learn more than shallow correlations between pairs of a sentence and visual features. That is, a particular dataset should not have a shortcut that only detecting nouns (object class) can perform well. Our dataset mitigates this issue by forcing instance-level recognition. We create a requirement that grounding the referring expressions must identify the target object among many distractors from the \textit{contrast set} (same class distractors). 

\begin{figure}[t] 
\begin{center}
   \includegraphics[width=1.00\linewidth]{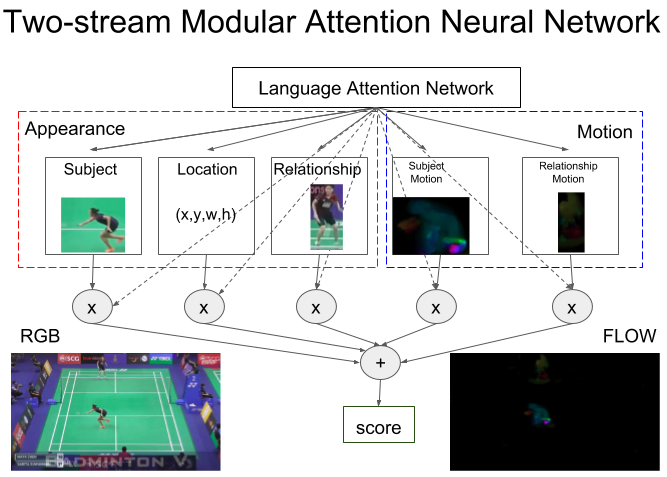}
\end{center}
   \caption{We add a motion stream to modular attention network. Our motion modules take optical flow input and model motion information for the subject and its relationship.}
   \label{fused1}
   \vspace{-1\baselineskip}
   \end{figure}


\begin{figure*}[t] 
\begin{center}
   \includegraphics[width=0.75\linewidth]{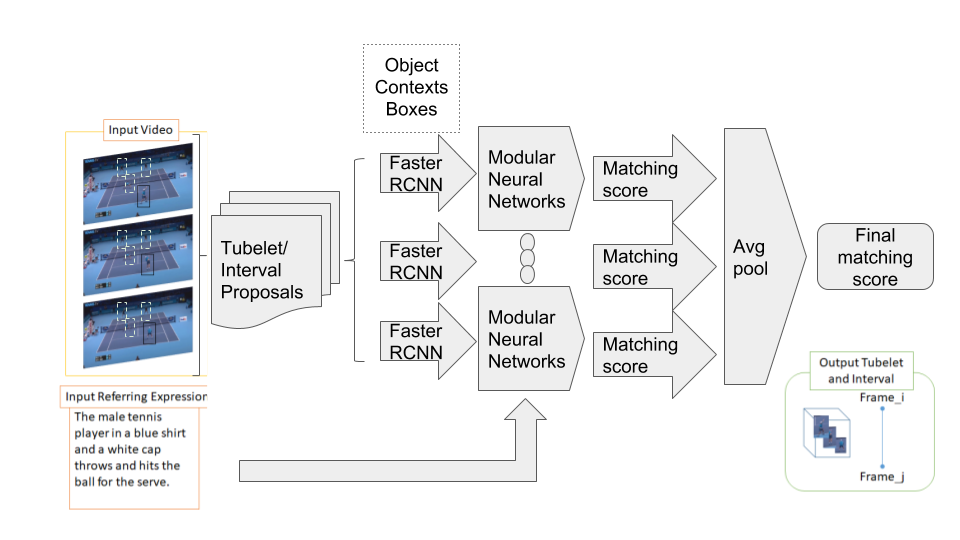}
\end{center}
\vspace{-1.5\baselineskip}
   \caption{An overview of our system: an input video is using either ground truth annotations or is fed into both tubelet object proposal and temporal interval proposal modules. The resulting tubelet and interval proposals are then fed into an appearance and motion Faster-RCNN to extract the two-stream features. Then, a modular neural network will rank the tubelets given an input referring expression. The scores are average-pooled, and the system outputs the most likely tubelet that contains the reference object.}
   \label{sys_overview}
\label{fig:long}
\label{fig:onecol}
\vspace{-1\baselineskip}
\end{figure*}

The contributions of this paper are (i) We propose a novel vision and language data collection scheme based on grammatical constraints for surface realization to ground video referring expressions in both space and time with lexical correlations between vision and language. We collected the \textbf{\textit{Spatio-Temporal Video Identifying Description Localization (STV-IDL)}} dataset consisting of 199 video sequences from Youtube and 7,569 identifying descriptions. (ii) We propose an interpretable system based on  \textit{\textbf{two-stream modular attention network}} that models both appearance and motion to ground referring expressions as instance-level video object detection and event localization. We also perform ablation studies to get insights and identify potential challenges for the task.

\section{Spatio-Temporal Localization}
Given ground truth temporal intervals ($[start, end]$) and object tubelets (a sequence of object bounding box coordinates in a given temporal interval, $\{[x_0, y_0, x_1, y_1]_{start}$,\ldots, $[x_0, y_0, x_1, y_1]_{end}\}$), we want to localize an identifying expression $ie$ to the correct target tubelet $tb_{target}$ not the distractor tubelets $tb_{distractor}$ as our predicted tubelet $r$. We evaluate using the accuracy measure.

For automatic localization, tubelet IoU \cite{ILSVRC15} and temporal IoU are used to evaluate the bounding box and temporal interval with the ground truth respectively. Let $R_i$ be the region in the frame $i$ to be detected,
\begin{equation}
tubelet\ IoU = \frac{\sum_{i}{\delta(IoU(r_i,R_i) > 0.5)}}{N},
\end{equation}
where the denominator is the number of detected frame measured by the standard Intersection over Union (IoU) in an image and $N$ denotes the number of union frames.
\begin{equation}
temporal\ IoU = \frac{\cap(interval_{i},interval_{j})}{\cup(interval_{i},interval_{j})},
\end{equation}
where $interval_{i}$ and $interval_{j}$ are input temporal intervals and the intersection and union functions are operations over 1-D intervals.

\section{Related Work}
\ \ \textbf{Spatio-Temporal Localization.} Spatio-temporal localization (or action understanding) is a long standing challenge in computer vision. Most existing datasets like LIRIS-HARL \cite{wolf2014evaluation}, J-HMDB \cite{jhuang2013towards}, UCF-Sports \cite{rodriguez2008action}, UCF-101 \cite{soomro2012ucf101} or AVA \cite{gu2018ava} localize a spatio-temporal tubelet for human actions in either trimmed videos or a simple visual setting or a fixed lexicon. In contrast to action labels, our work accepts a free-form referring expression annotation which also contains a richer set of relations in the forms of prepositions, adverbs and conjunctions.

\textbf{Referring Expression Comprehension.}
The goal of referring expression comprehension \cite{golland2010game} is to ground phrases or sentences into the specific visual regions that the phrase refers. Prior works in the image domain have either focused on using a captioning module to generate the sentence \cite{mao2016generation, nagaraja2016modeling} or learning a joint embedding to comprehend the sentence by modeling the corresponding region unambiguously and localize the region \cite{rohrbach2016grounding, wang2016learning, hu2017modeling, yu2018mattnet}.   For the video domain, \cite{yamaguchi2017spatio} further annotated the ActivityNet dataset with one referring expression per video for video retrieval with natural language query. \cite{li2017tracking} uses referring expressions to help track a target object in a video sequence in a subset of OTB100 \cite{lu2014online} and ImageNet VID \cite{ILSVRC15}. DiDeMo \cite{hendricks17iccv} and TEMPO \cite{hendricks2018localizing} focus on localizing an input sentence into the corresponding temporal interval out of a finite number of backgrounds. Importantly, these datasets do not consider distractor objects from the same class. While our work also focuses on the video domain, it focuses on localizing objects and events as spatio-temporal tubelets aligned with an input expression.

\textbf{Surface Realization in Vision and Language.} Surface realization is a process for generating surface forms, like natural language sentences, based on some underlying representations. For natural language generation, the underlying representation tends to be syntactic features. In vision and language, captioning systems can use meaning representation like triplets as an input for a surface realization module to generate a sentence. \cite{farhadi2010every} uses \textless Objects, Actions, Scenes\textgreater. \cite{yang2011corpus} uses part-of-speech as \textless Nouns, Verbs, Scenes, Prepositions\textgreater. \cite{li2011composing} uses \textless \textless adj1, obj1\textgreater, prep, \textless adj2, obj2\textgreater \textgreater where adjectives are object attributes and prepositions are spatial relationships between objects. TEMPO \cite{hendricks2018localizing} and TVQA \cite{lei2018tvqa} use a compositional format for words like before or after to specify temporal relationships between events during crowdsourcing. 

 We incorporate grammatical constraints (Linguistic prescription) based on part-of-speech into our annotation pipeline so that we can crowdsource well-formed sentences from people which contain enough meaning representations for vision systems to locate the target object with visual contexts. Instead of manually writing sentences based on context-free grammars like \cite{yu2013grounded}, we ask the annotators to write sentences in which valid sentences contain at least a noun phrase (NP), a verb phrase (VP) and one of a prepositional phrase (PP), adverb phrase (ADVP) or conjunction phrase (CONJP). The rest of each sentence are language variations where we expect crowdsourcing to create more variations compared to manual annotations by a few annotators. We want computer vision models to learn useful and interpretable features by correlating the expressions and videos. So, we want the learned visual semantics from grounding models to be similar to structural inputs in surface realization systems. Each part-of-speech correlates with a specific visual feature. 

\section{STV-IDL Dataset}
\subsection{Dataset Construction}
We develop a new data collection schema that ensures rich correspondences between referring expressions and referred objects in a video using constraints. The spatio-temporal relations that we are interested in are about state transitions, that is, what happens before and after the action and how objects move. The state transitions should be relative to other objects and background. For example, a sentence \textit{`A man in a green uniform kicking the ball then running toward the net.'} is a good video referring expression. This sentence is valid only in a spatial region that represents a noun phrase \textit{`a man in a green uniform'} and a time interval in which an action from the verb phrase \textit{`hitting the ball then running toward the net'} occurs. Also, the action \textit{`hitting the ball'} comes before his next action \textit{`running toward the net'} which shows the action steps of \textit{`hitting'} followed by \textit{`running'} and the action \textit{`running'} has a context object \textit{`the net.'}

\begin{table}[] \label{table_dataset1}
\centering
\caption{STV-IDL dataset statistics.}
\label{num_sent}
\begin{tabular}{lcc} 
\HeaderColor Info    &  Statistics  \\
\hline
\RowColor Number of Videos & 199  \\
Number of Sentences & 7569  \\
\RowColor Average objects per Video & 2.85  \\
Average words per Sentence & 22.65 \\
\RowColor Sentences per Video & 38.04 \\
\hline
\end{tabular}
\end{table}

First, we ensure that all of our High Definition videos (720p) crawled from Youtube contain at least two objects similar to \cite{mao2016generation}, but each video will focus on just one object class to form a contrast set. This constraint prevents a simple video object detector from resolving referential ambiguity using only nouns by just outputting based on class information as in \cite{cirik2018visual}. Because a simple object detector randomly outputs one object from a combination of the target and the contrast set. Then, the output is the same as random because the object confidence scores do not correlate with the referring expression. We want more language cues to guide the system to seek additional visual contexts \cite{divvala2009empirical} to focus and output \textit{only one} unambiguous object detection. The dataset contains 13 categories of videos which are either multi-player sports or animals. Second, inspired by \cite{siskind1990acquiring, yu2013grounded}, a sentence, consists of a subject and a predicate, can be viewed as a set of structured labels based on part-of-speech and each label can be meaningfully grounded in a video. Besides, annotations can use \textit{grammars for lexical grounding and surface realization}. Therefore, we ensure that every referring expression in our dataset provides grammatically relevant visual grounding based on part-of-speech such that a valid sentence must contain at least a noun phrase (NP), a verb phrase (VP) and one of a prepositional phrase (PP), adverb phrase (ADVP) or conjunction phrase (CONJP). We also found that the annotators may write relevant sentences without the constraints but the contents are random and may not be visually grounded either spatially or temporally or both in the video. Some example sentences without the constraints are \textit{``The guy was lucky to save the tennis ball."} and \textit{``The sun is blocking the ball for the back player."}.

\begin{table}[] \label{table_dataset2}
\centering
\caption{STV-IDL part-of-speech statistics. (Please see the supplementary material for more details.)}
\label{num_pos}
\begin{tabular}{lcc} 
\HeaderColor Part-of-Speech    &  percents  \\
\hline
\RowColor Noun, singular or mass (NN) & 28.1  \\
Determiner (DT) & 15.3 \\
\RowColor Preposition or 
 & 10.9 \\ 
 \RowColor subordinating conjunction (IN) & \\
 Adjective (JJ) & 9.7 \\ 
\RowColor Possessive pronoun (\$PRP) & 5.7 \\
Verb, 3rd person  & 5.1 \\
singular present (VBZ) & \\
\RowColor Adverbs (RB) & 3.6   \\
Coordinating conjunction (CC) & 3.4 \\  
 
\hline
\end{tabular}
\end{table}

\subsection{Video Tubelet, Temporal Interval, and Expressions Annotations}
We manually identify interesting events in each video and select a keyframe for that action in the presence of distractors. Then, we manually annotate the start and end of that event into an interval lasting around one second. For bounding box annotation, we use a javascript variant of Vatic \cite{vondrick2013efficiently, vaticjs} to manually draw a tubelet of bounding boxes in all frames for each object of interest in every video. We crowdsource annotations of referring expressions from Amazon Mechanical Turk (AMT). We create a clip segment with a bounding box around the target object to fixate the annotator's attention. 

Next, we manually verify the referring expressions using another web interface that helps us evaluate if the sentence refers to the target object, is correct based on the video, is different from sentences for the distractors and is sufficient to distinguish the target object from the distractors and the background. The annotation interfaces, payment and dataset statistics are shown in supplementary material. We refer to the resulting referring expressions as \textit{identifying descriptions} \cite{mitchell2013generating} because our expressions are referring expressions in the verified intervals which may be overspecified but are also descriptions which may be underspecified for the whole videos. Our referring expressions are long because we want to make sure that they are clear enough to provide input cues for the system. However, it still might be not enough to localize an event from the whole video because the video has many events and can be exhaustive to be specific for a particular event.

\begin{figure}[t] 
\begin{center}
   \includegraphics[width=1.00\linewidth]{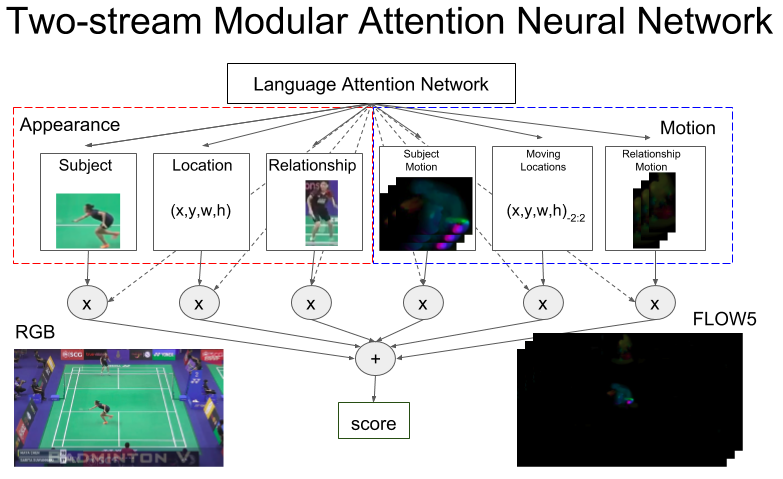}
\end{center}
   \caption{Stacked two-stream modular attention network based on five optical flow image input. We model the bounding box sequence is a moving location module and a relationship module. The motion Faster-RCNN is also trained using a stack of five flow images for frame index $f_i \in [t-2, t+2]$.}
   \label{fig2a}
\end{figure}

\section{Approach: Two-stream Modular Attention Network}
We start by employing a state-of-the-art image referring expression localization, namely, Modular Attention Network (MAttNet) \cite{yu2018mattnet} for our tasks. This model fits our objective since it is a variant of modular neural networks \cite{auda1999modular,andreas2016neural} that is decomposed based on tasks according to Fodor's modularity of mind \cite{fodor1985precis}. Therefore, we can interpret the model in an ablation study on each neural module for a specific vision subtask and input type. Also, the model also provides linguistic interpretability using its language attention module that can visualize different bindings from word symbols in a referring expression to each visual module as attention $a_{m,t}$ where module $m \in \{subj, loc, rel\}$ (subject, location, relationship) and $t$ is the index location of the word this attention weights its hidden representation the Bi-LSTM encodes. 

The original MAttNet model (RGB) decomposes image referring expression grounding into three modules, a subject module, a location module, and a relationship module. The network output score for an object $o_i$ and an expression $r$ is,
\begin{equation}
S(o_i|r) = \sum_{m \in modules}{w_m S(o_i|q^m)},
\end{equation}
where $w_m$ is the weight vector from the language attention module on the visual module $m$. $q^m$ is the weighted sum of attention $a_{m,t}$ over all word embedding. $S(o_i|q^m)$ is the module score from a cosine similarity in the joint embedding between the visual representation of $o_i$ denoted as $\widetilde{v_i}^m$ and $q^m$.

Given a positive pair $(o_i, r_i)$, the network is discriminatively trained by sampling two negative pairs $(o_i, r_j)$ and $(o_k, r_i)$ where $r_j$ is the expression from other contrast object and $o_k$ is the contrast object from the same frame. The combined hinged loss $L_r$ is,
\begin{multline}
L_r = \sum_{i}{\lambda_1 \max(0, \Delta + S(o_i,r_j) - S(o_i, r_i))} \\ + \lambda_2 \max(0, \Delta + S(o_k,r_i) - S(o_i, r_i)).
\end{multline}
The loss is linearly combined with other loss terms such as attribute prediction with cross-entropy loss $L_{att}$ from the subject module in a multi-task learning setting.

We extend MAttNet to the video domain by applying two things. First, MAttNet uses Faster-RCNN \cite{girshick2015fast} for feature extraction so we follow a well-established actor-action detection pipeline which extends image object detection to frame-based spatio-temporal action detection \cite{peng2016multi}. With this, we reframe the problem by replacing action labels with referring expressions and putting MAttNet on top of Faster-RCNN. Also, we use external object and interval proposal instead of Region Proposal Network (RPN) in Faster-RCNN. Second, we add subject motion and relationship motion modules to capture temporal information in a two-streams setting \cite{simonyan2014two}. These modules have the same architecture as the subject and relationship module but are using optical flow as their input. We replace the three channel RGB input with a stack of flow-x, flow-y and flow magnitude from the flow image. The aim of these modifications, depicted in Figure \ref{fused1}, is to better model attributes, motion, movements and dynamic context in a video.

Previous work \cite{simonyan2014two} has shown that stacking many optical flow images can help recognition. So, we train another variant of two-stream modular attention network using stacked five optical flow frames shown in Figure \ref{fig2a}. In this setting, we train the stacked motion Faster-RCNN by stacking flow images $F_{idx}$ where frame index ${idx} \in [t-2, t+2]$. The input becomes a 15 channel stacked optical flow image. In addition, we add the moving location module to further model the movement of the location by stacking location features $l_i = [\frac{x_{min}}{W},\frac{y_{min}}{H},\frac{x_{max}}{W},\frac{y_{max}}{H},\frac{Area_{region}}{Area_{image}} ]$ where \textit{W} and \textit{H} are width and height of the image. Then the location features are concatenated with the location difference feature of the target object with up to five context objects from the same class, $\delta_{ij} = [\frac{\Delta x_{min}}{W},\frac{\Delta y_{min}}{H},\frac{\Delta x_{max}}{W},\frac{\Delta y_{max}}{H},\frac{\Delta Area_{region}}{Area_{image}} ]$ so that we have a sequence of $[l_i;\delta_{ij}]_{idx}$ where frame index ${idx} \in [t-2, t+2]$. Then, we place an LSTM on top of the sequence and we forward the concatenation of all hidden states to a fully connected layer and output the final location features. We also make a location sequence and place an LSTM on top of location in the relationship motion module in this stacked optical flow setting.

\subsection{Tubelet and Temporal Interval Proposals}
We employ the state-of-the-art video object detector, flow-guided feature aggregation (FGFA) \cite{zhu2017flow}, finetuned on STV-IDL to generate the tubelet proposals. The per-frame detections from FGFA are post-processed by linking into tubelets using Seq-NMS \cite{han2016seq} based on the top 300 bounding boxes ranked by the confidence of the category scores.

For temporal proposals, we implemented a varient of Deep Action Proposals (DAPs) \cite{escorcia2016daps} based on multi-scale proposal. First, we use a temporal sliding window with a fixed length of \textit{L} frames and a stride of \textit{s} (8 in our case). This produces a set of intervals, \textit{($b_i$,$e_i$)} where $b_i$ and $e_i$ are the beginning and the end of the interval. Then, we extract the C3D features \cite{tran2015learning} from the image frames in that interval using the activation in the `fc7' layer, pretrained on the Sports-1M dataset \cite{KarpathyCVPR14}. The feature set $f = C3D(t_i:t_i+\delta),$ $ t_i \in [b_i,e_i]$ where $\delta=16$ from the original pretrained model. The duration of each segment $L_k$ also increases as a power of 2, that is $L_{k+1}=2*L_k$. The features are fed to a 2-layered LSTM to perform \textit{\{Event/Background\}} sequence classification.

\begin{table}[] \label{task1}
\centering
\caption{Identifying Description Localization: mAP for each collection. (values are in percents.) The fused1 MAttNet is the proposed two-stream method and the fused5 MAttNet is the stacked version of the proposed two-stream method.}
\label{task1}
\begin{tabular}{lc}
\HeaderColor Model    & mAP  \\
\hline
\RowColor random & 29.68 \\
RGB MAttNet & 41.51  \\
\hline
\RowColor flow MAttNet & 39.02 \\
flow5 MAttNet & 41.90  \\ 
\hline
\RowColor fused1 MAttNet &  \textbf{44.66}   \\
fused5 MAttNet & 42.82  \\ 
\hline
\end{tabular}
\end{table}

\begin{table}[] \label{abla1}
\centering
\caption{Ablation study on fused1 MAttNet: mAP for each module combination. (values are in percents.)}
\label{abla1}
\begin{tabular}{lc}
\HeaderColor Model    & mAP \\
\hline
\RowColor Subject+Location & 44.46  \\
+Relationship & 44.46 \\
\RowColor +Subject Motion & 44.46  \\ 
+Relationship Motion &  \textbf{44.66}   \\
\hline
\end{tabular}
\vspace{-1\baselineskip}
\end{table}

\section{Experiments and Analysis}
We want to show how and to what extent modular attention networks ground input expressions with motion information in videos. So, we perform two sets of experiments, identifying description localization and automatic video object detector and temporal event localization. Similar to \cite{gu2018ava}, we split the dataset into training, validation and test sets at the video level; that is, there are no overlapping video segments for every split. There are 159 training, 13 validation, and 27 test videos. The rough ratio is 12:1:2. Implementation details are in the supplementary material.

\subsection{Identifying Description Localization}
\textbf{Setup.} We perform three experiments, localization with ground truth annotations, module ablation study, and word attention study. First, we evaluate our model by selecting the target from a pool of candidate targets plus distractors. We compare five models based on input and modules. The five models are (1) MAttNet for RGB input (RGB MAttNet/original model/baseline); (2) MAttNet for flow image input (flow MAttNet); (3) MAttNet for stacked five flow image input (flow5 MAttNet); (4) two-stream MAttNet for RGB and flow image input (fused1 MAttNet) and (5) two-stream MAttNet for RGB and stacked five flow image input (fused5 MAttNet). Second, we interpret the model by setting the module score weights from language attention module to zeros for the modules we want to turn off in our ablation study. Third, we collect the statistics of the attention of each word from the input expressions in the test set to explain how and which kind of words each module attends. 

\begin{figure*}
    \centering
    \begin{subfigure}[b]{0.3\textwidth}
        \includegraphics[width=\textwidth]{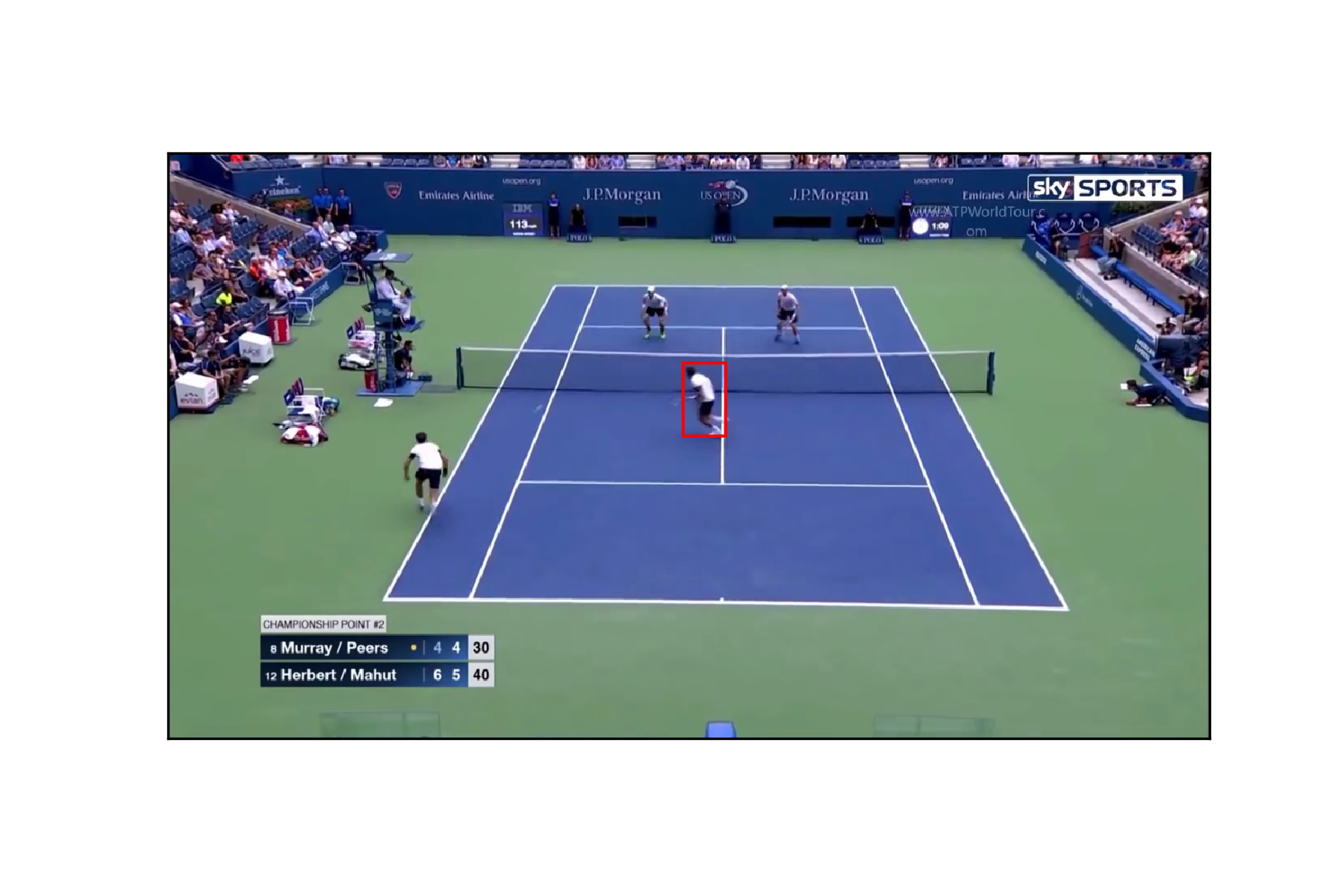}
        \label{fig:gull}
    \end{subfigure}
    ~ 
    \begin{subfigure}[b]{0.3\textwidth}
        \includegraphics[width=\textwidth]{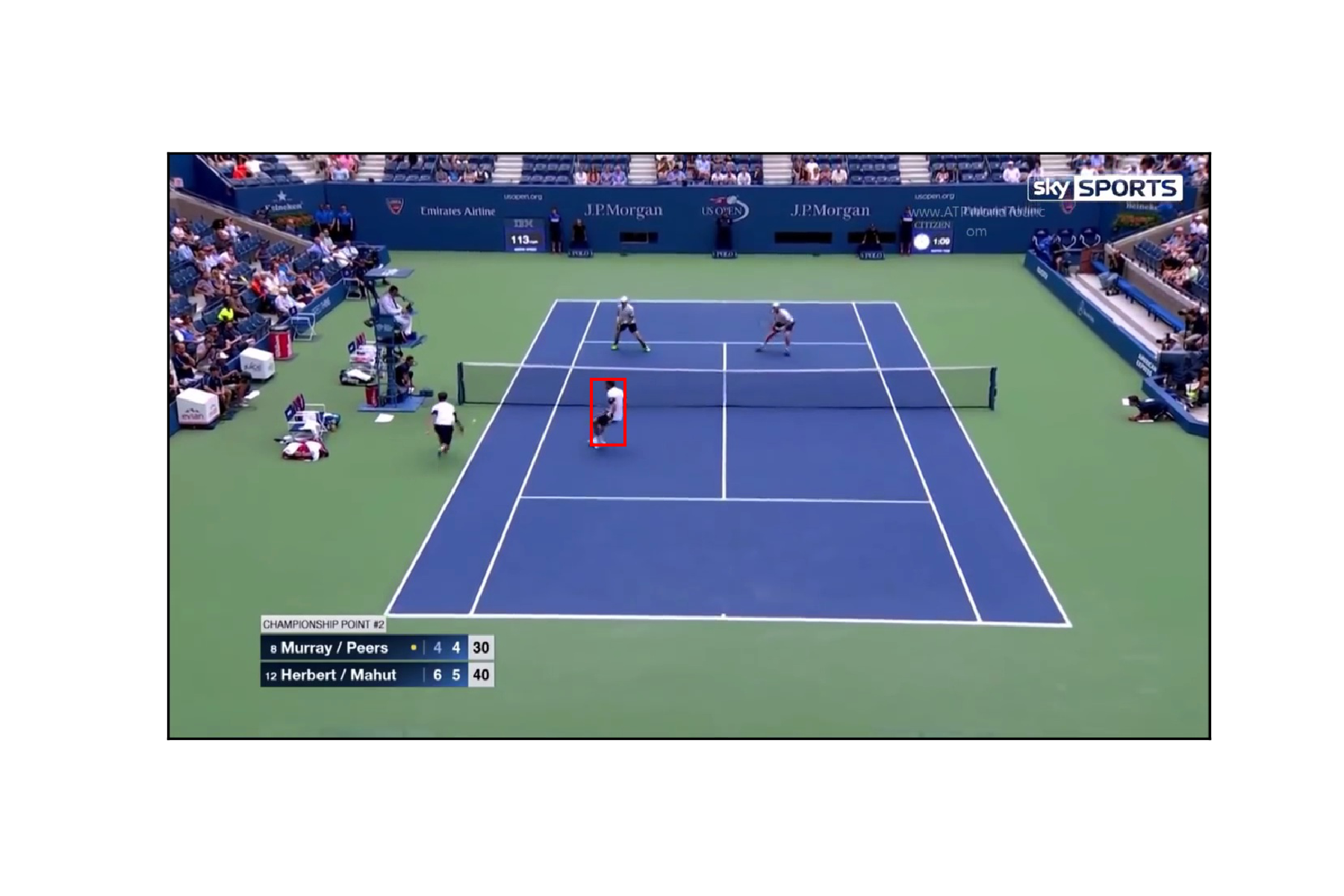}
        \label{fig:tiger}
    \end{subfigure}
    ~ 
    \begin{subfigure}[b]{0.3\textwidth}
        \includegraphics[width=\textwidth]{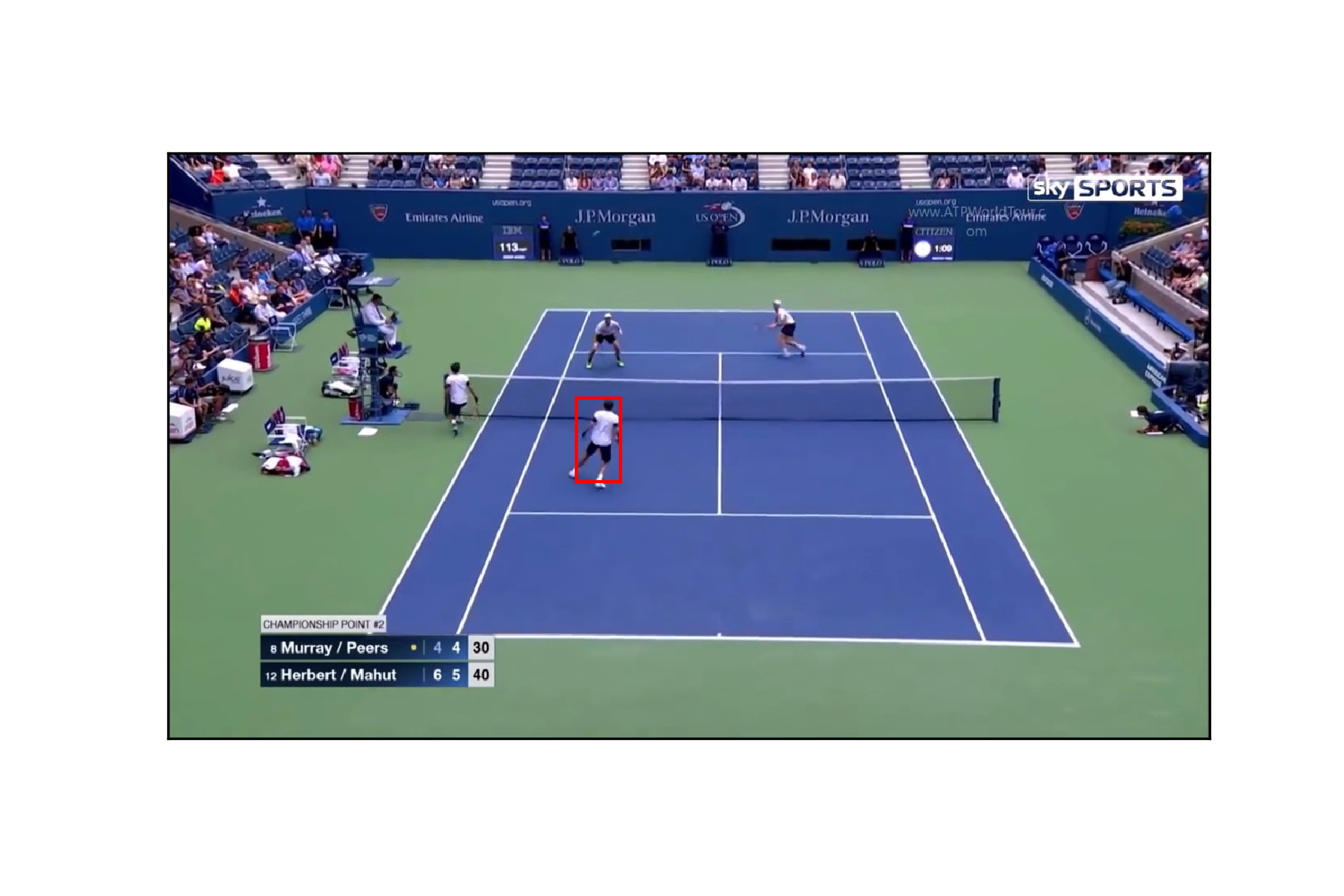}
        \label{fig:mouse}
    \end{subfigure}
    \vspace{-2\baselineskip}
    \caption{A qualitative result: the first, middle and last frames from an interval in the STV-IDL dataset with an expression, `The male tennis player in the near court moves from right to left in order to hit the ball but his teammate outside the court reaches the ball first and just hits it.' The fused1 MAttNet can properly refer to the object highlighted in the red box in contrast to the baseline.}
    \label{fig:qualitative}
\end{figure*}

\begin{figure*}
    \centering
    \begin{subfigure}[b]{0.98\textwidth}
        \includegraphics[width=\textwidth]{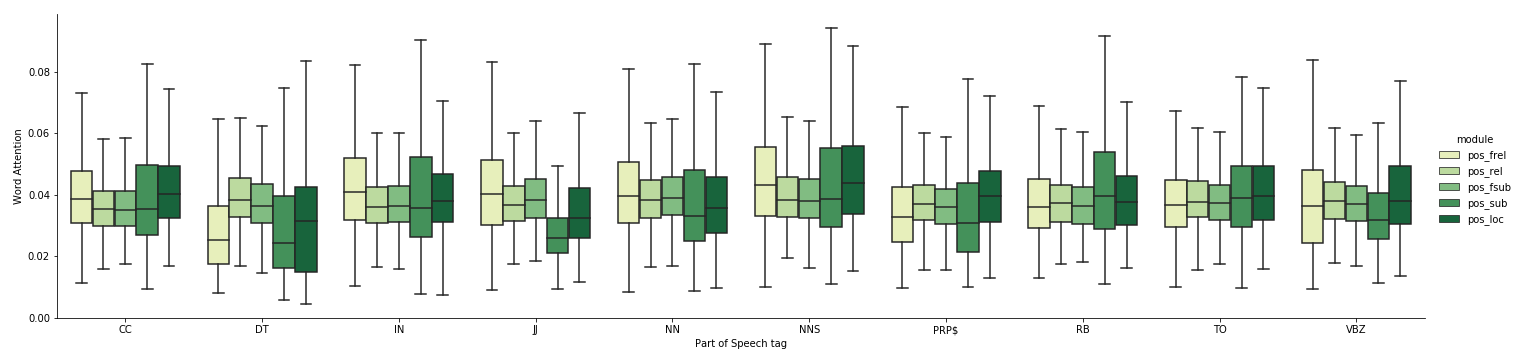}
        \label{fig:gull}
        \vspace{-1\baselineskip}
    \end{subfigure}
    \caption{Aggregations of output word attention weights for each module on the STV-IDL test set. Part-of-speech tags are CC, DT, IN, JJ, NN, NNS, PRP\$, RB, TO and VBZ (left to right).}
    \label{fig:word_attention}
     \vspace{-1\baselineskip}
\end{figure*}

\begin{table}[] \label{abla5}
\centering
\caption{Ablation study on fused5 MAttNet: mAP for each module combination. (values are in percents.)}
\label{abla5}
\begin{tabular}{lc}
\HeaderColor Model    & mAP  \\
\hline
\RowColor Subject+Location & 33.97  \\
+Relationship & 35.32 \\
\RowColor +Subject Motion & 35.41  \\ 
+Moving Location &  \textbf{42.84}   \\
\RowColor +Relationship Motion & 42.82  \\ 
\hline
\end{tabular}
 \vspace{-1\baselineskip}
\end{table}


\begin{figure}[t] 
\begin{center}
   \includegraphics[width=0.75\linewidth]{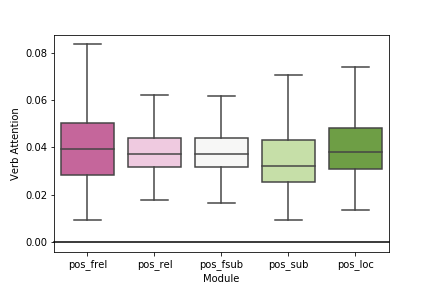}
    \vspace{-1\baselineskip}
\end{center}
   \caption{Aggregations on all verbs for each module. (from left to right: Relationship flow/RGB, Subject flow/RGB, Location)}
   \label{wa_verbs} 
 \vspace{-1\baselineskip}
\end{figure}
\textbf{Results.} The accuracies in Table \ref{task1} show superior performance for stacked flow5 and two-streams models. The stacked flow5 model improves over the RGB baseline by $0.39\%$ while two-stream fused1 and fused5 models have $3.15\%$ and $1.31\%$ improvement respectively. Both variants of two-stream models, fused1 and fused5, outperform all one-stream models, RGB, flow, and flow5. All models perform better than randomly selecting an object from the set of tubelets. 

The accuracies in Table \ref{abla1} show that each module in fused1 learns better since the modules in appearance stream alone have $2.95\%$ improvements over the RGB only baseline. We further hypothesize that \textit{the reason is the motion stream takes care of motion grounding so the appearance modules can learn better because of the separation of unrelated information into other modules. A modular neural network avoids internal interference between features by training each module independently and each module will masters its task more precisely \cite{auda1999modular}.} The additional relationship motion module also provides complementary information for the additional $0.20\%$ improvement. 
The accuracies in Table \ref{abla5} show that the stacked flow5 model focuses mostly on the moving location module which causes the overall improvement over the RGB baseline. \textit{The moving location is a predictive feature to model motion and spatial location \cite{yin2017obj2text}, but it prevents other vision modules from becoming sufficiently tuned in this setting.} We also try to combine the moving location with the fused1 setting. The results degenerate more, and the overall accuracy is only $37.12\%$. It is even lower than flow MAttNet model.

Figure \ref{fig:word_attention} shows how the language attention network assigns weights to each module by aggregating all the weights for each word based on Penn part-of-speech tag during test set prediction of the fused1 model to explain the performance gain. The aggregated statistics show that motion words like verbs, prepositions, and conjunctions are ranked higher for flow modules on average which means more attention to motion. We also focus on just aggregating verbs in Figure \ref{wa_verbs} to further explain the modules. The statistics show that flow and location modules focus more on verbs on average compared to their corresponding appearance-based modules.

\subsection{Automatic Video Object Detector and Temporal Event Localization}
Because spatio-temporal detection and localization is very challenging, we want to identify potential challenges for spatio-temporal grounding when automatic computer vision systems replace the ground truth annotations. So, we replace tubelets with top $8$ detections from flow-guided feature aggregation (FGFA) \cite{zhu2017flow} and temporal intervals with the proposal system described in Section 3.2. We create three scenarios: in each scenario, varying amounts of the problem are revealed via the ground truth to separate each component and measure the hardness of each subproblem and the impact of one on another.
\subsubsection{Automatic Video Object Detector}
\textbf{Setup.} We evaluate both the tubelet object proposals and the pretrained modular attention networks. We replace the groundtruth tubelets to imperfect proposals which contains bounding box perturbations and we want to see how the model behaves.

\textbf{Results.} Since all modular attention networks are not trained on tubelet proposals, the results from the automatic video object detector in Table \ref{task2} shows performance drops in all models and the performances are even lower than the object detection baseline. The object detection baseline selects the tubelet with the highest confidence score from FGFA. We hypothesize that it is from bounding box perturbation that may affect both Faster RCNN features and location features. The results also show that the performance drops are more severe in two-stream models - we think that it is from an accumulation of errors from both streams. 

\begin{table}[] 
\centering
\caption{Visual Object Detection: mAP tracklet IoU@0.5 for each model. (values are in percents.)}
\label{task2}
\begin{tabular}{lc}
\HeaderColor Model    & IoU@0.5  \\
\hline
\RowColor RGB MAttNet & \textbf{35.02}   \\
flow MAttNet & 22.63 \\
\RowColor flow5 MAttNet & 28.98  \\ 
fused1 MAttNet &  23.93   \\
\RowColor fused5 MAttNet & 24.26 \\ 
\hline
\textit{FGFA most conf.} & \textit{35.87} \\
\RowColor \textit{FGFA 2nd conf.} & \textit{34.16} \\
\hline
\end{tabular}
 \vspace{-1\baselineskip}
\end{table}

\begin{table}[] 
\centering
\caption{Event Localization: mAP temporal IoU@0.5 for each model. (values are in percents.)}
\label{task3}
\begin{tabular}{lc}
\HeaderColor Model    & tIoU@0.5  \\
\hline
\RowColor RGB MAttNet & 8.72   \\
flow MAttNet & 7.28 \\
\RowColor flow5 MAttNet & \textbf{8.79}  \\ 
fused1 MAttNet &  8.07   \\
\RowColor fused5 MAttNet & 7.02 \\ 
\hline
 speaker LSTM & 7.74 \\ 
\RowColor speaker Bi-LSTM & \textbf{10.10} \\ 
\hline
\end{tabular}
\vspace{-1\baselineskip}
\end{table}

\subsubsection{Temporal Event Localization}
\textbf{Setup.} We evaluate the event localization component by removing ground truth temporal intervals. All previous settings so far operate on trimmed video segments and focus on `where' the sentences refer to. We want to see how the model behaves on untrimmed videos in which the system needs to answer `when' the referred events occur. The system's task is to infer the temporal intervals $[t_k, t_{k+40})$ which are likely to correspond to the input expressions. We evaluate the system via temporal mean Average Precision with temporal IoU similar to \cite{krishna2017visual}. Since our identifying descriptions are sentences for the whole videos, we compare modular attention networks to a video captioner, S2VT \cite{venugopalan15iccv}, which is a speaker model \cite{mao2016generation} that output the probability of producing an expression given a video. The S2VT model is trained on a different feature set consisting of the image features from the last layer `fc1000' of ResNet-50 \cite{he2016deep}, the interval \textit{($b_i$,$e_i$)} and the current frame number. This S2VT model is trained on ground truth intervals and expressions, so it is likely to produce expressions with high probabilities on the ground truth event intervals compared to the background intervals which do not contain `interesting' events.

\textbf{Results.} The results in Table \ref{task3} shows that speaker Bi-LSTM performs the best and even better than all modular attention networks. We suspect that the reason is from the discriminative training scheme of the modular attention networks is not suitable for temporal localization. Training with only negative pairs from the same frame takes a week, so it is computationally expensive to train with all negative pairs from all frames in the whole video. The top-5 prediction for Bi-LSTM increases to $26.23\%$ but it is still far from the upper bound of $71.02\%$, the recall of the proposal system. 

\subsubsection{Spatio-temporal Localization}
\textbf{Setup.} We evaluate our event interval proposals, tubelet object proposals, and modular attention networks.  We fix tubelet Intersection over Union (tubelet IoU) to 0.5. The evaluation is a two-step process, temporal IoU then tubelet IoU. We allow tubelet IoU over all frames of the proposal interval instead of ground truth interval to show that the system refers to the right object in an event interval and the tubelet IoU does not depend on temporal IoU. 

\textbf{Results.} The results in Table \ref{task4} show that the performance further decreases from Table \ref{task3}. We suspect that the reason is also from the discriminative training scheme because the models are not trained on some background frames.

\begin{table}[] 
\centering
\caption{Spatio-temporal Localization: mAP temporal IoU@0.5 then tracklet IoU@0.5 for each model. (values are in percents.)}
\label{task4}
\begin{tabular}{lc}
\HeaderColor Model    & tIoU@0.5  \\
\hline
\RowColor RGB MAttNet & \textbf{2.75}   \\
flow MAttNet & 2.04 \\
\RowColor flow5 MAttNet & 2.62  \\ 
fused1 MAttNet & 1.70   \\
\RowColor fused5 MAttNet & 1.51 \\ 
\hline
\end{tabular}
 \vspace{-1\baselineskip}
\end{table}

\section{Summary}
We discussed the problem of grounding spatio-temporal identifying descriptions to spatio-temporal object-event tubelets in videos. The critical challenge in this dataset is to ground verbs and motion words in both space and time, and we show that this is possible by our proposed two-stream modular neural network models which have complimentary optical flow inputs to ground verbs and motion words. We validate this by collecting aggregated statistics on word attention and found that the two-stream models ground verbs better. The motion stream also helps the appearance stream learn better because it abstracts away motion noise from appearance. We further inspected the components in the system and revealed potential challenges. A better training scheme such as improved loss functions or hard example mining for future spatio-temporal grounding systems should consider both efficiency and effectiveness.

\section{Acknowledgement}
The authors thank the anonymous reviewers for their insightful comments and suggestions. We thank Dr. Hal Daum\'e III, Nelson Padua-Perez and Dr. Ryan Farrell for very useful advises and discussions. We also thank members of UMD Computer Vision Lab (CVL), UMD Human-Computer Interaction Lab (HCIL) and UMD Computational Linguistics and Information Processing Lab (CLIP) for useful insights and support. We also thank academic twitter users for useful knowledge and discussions.

\bibliography{naaclhlt2019}

\begin{thebibliography}{53}
\expandafter\ifx\csname natexlab\endcsname\relax\def\natexlab#1{#1}\fi

\bibitem[{Andreas et~al.(2016)Andreas, Rohrbach, Darrell, and
  Klein}]{andreas2016neural}
Jacob Andreas, Marcus Rohrbach, Trevor Darrell, and Dan Klein. 2016.
\newblock Neural module networks.
\newblock In \emph{Proceedings of the IEEE Conference on Computer Vision and
  Pattern Recognition}, pages 39--48.

\bibitem[{Auda and Kamel(1999)}]{auda1999modular}
Gasser Auda and Mohamed Kamel. 1999.
\newblock Modular neural networks: a survey.
\newblock \emph{International Journal of Neural Systems}, 9(02):129--151.

\bibitem[{Berzak et~al.(2015)Berzak, Barbu, Harari, Katz, and
  Ullman}]{berzak2015you}
Yevgeni Berzak, Andrei Barbu, Daniel Harari, Boris Katz, and Shimon Ullman.
  2015.
\newblock Do you see what i mean? visual resolution of linguistic ambiguities.
\newblock In \emph{Proceedings of the 2015 Conference on Empirical Methods in
  Natural Language Processing}, pages 1477--1487.

\bibitem[{Bolkensteyn()}]{vaticjs}
Dinesh Bolkensteyn.
\newblock vatic.js: A pure javascript video annotation tool.
\newblock \url{https://dbolkensteyn.github.io/vatic.js/}.

\bibitem[{Cirik et~al.(2018)Cirik, Morency, and
  Berg-Kirkpatrick}]{cirik2018visual}
Volkan Cirik, Louis-Philippe Morency, and Taylor Berg-Kirkpatrick. 2018.
\newblock Visual referring expression recognition: What do systems actually
  learn?
\newblock In \emph{Proceedings of the 2018 Conference of the North American
  Chapter of the Association for Computational Linguistics: Human Language
  Technologies, Volume 2 (Short Papers)}, volume~2, pages 781--787.

\bibitem[{Dale and Reiter(1995)}]{dale1995computational}
Robert Dale and Ehud Reiter. 1995.
\newblock Computational interpretations of the gricean maxims in the generation
  of referring expressions.
\newblock \emph{Cognitive science}, 19(2):233--263.

\bibitem[{Divvala et~al.(2009)Divvala, Hoiem, Hays, Efros, and
  Hebert}]{divvala2009empirical}
Santosh~K Divvala, Derek Hoiem, James~H Hays, Alexei~A Efros, and Martial
  Hebert. 2009.
\newblock An empirical study of context in object detection.
\newblock In \emph{Computer Vision and Pattern Recognition, 2009. CVPR 2009.
  IEEE Conference on}, pages 1271--1278. IEEE.

\bibitem[{Escorcia et~al.(2016)Escorcia, Heilbron, Niebles, and
  Ghanem}]{escorcia2016daps}
Victor Escorcia, Fabian~Caba Heilbron, Juan~Carlos Niebles, and Bernard Ghanem.
  2016.
\newblock Daps: Deep action proposals for action understanding.
\newblock In \emph{European Conference on Computer Vision}, pages 768--784.
  Springer.

\bibitem[{Farhadi et~al.(2010)Farhadi, Hejrati, Sadeghi, Young, Rashtchian,
  Hockenmaier, and Forsyth}]{farhadi2010every}
Ali Farhadi, Mohsen Hejrati, Mohammad~Amin Sadeghi, Peter Young, Cyrus
  Rashtchian, Julia Hockenmaier, and David Forsyth. 2010.
\newblock Every picture tells a story: Generating sentences from images.
\newblock In \emph{European conference on computer vision}, pages 15--29.
  Springer.

\bibitem[{Fodor(1985)}]{fodor1985precis}
Jerry~A Fodor. 1985.
\newblock Precis of the modularity of mind.
\newblock \emph{Behavioral and brain sciences}, 8(1):1--5.

\bibitem[{Gao et~al.(2017)Gao, Sun, Yang, and Nevatia}]{gao2017tall}
Jiyang Gao, Chen Sun, Zhenheng Yang, and Ram Nevatia. 2017.
\newblock Tall: Temporal activity localization via language query.
\newblock In \emph{Proceedings of the IEEE International Conference on Computer
  Vision}, pages 5267--5275.

\bibitem[{Gavrilyuk et~al.(2018)Gavrilyuk, Ghodrati, Li, and
  Snoek}]{gavrilyuk2018actor}
Kirill Gavrilyuk, Amir Ghodrati, Zhenyang Li, and Cees~GM Snoek. 2018.
\newblock Actor and action video segmentation from a sentence.
\newblock In \emph{Proceedings of the IEEE Conference on Computer Vision and
  Pattern Recognition}, pages 5958--5966.

\bibitem[{Girshick(2015)}]{girshick2015fast}
Ross Girshick. 2015.
\newblock Fast r-cnn.
\newblock In \emph{Proceedings of the IEEE international conference on computer
  vision}, pages 1440--1448.

\bibitem[{Golland et~al.(2010)Golland, Liang, and Klein}]{golland2010game}
Dave Golland, Percy Liang, and Dan Klein. 2010.
\newblock A game-theoretic approach to generating spatial descriptions.
\newblock In \emph{Proceedings of the 2010 conference on empirical methods in
  natural language processing}, pages 410--419. Association for Computational
  Linguistics.

\bibitem[{Gu et~al.(2018)Gu, Sun, Ross, Vondrick, Pantofaru, Li,
  Vijayanarasimhan, Toderici, Ricco, Sukthankar et~al.}]{gu2018ava}
Chunhui Gu, Chen Sun, David Ross, Carl Vondrick, Caroline Pantofaru, Yeqing Li,
  Sudheendra Vijayanarasimhan, George Toderici, Susanna Ricco, Rahul
  Sukthankar, et~al. 2018.
\newblock Ava: A video dataset of spatio-temporally localized atomic visual
  actions.
\newblock In \emph{CVPR 2018}.

\bibitem[{Han et~al.(2016)Han, Khorrami, Paine, Ramachandran, Babaeizadeh, Shi,
  Li, Yan, and Huang}]{han2016seq}
Wei Han, Pooya Khorrami, Tom~Le Paine, Prajit Ramachandran, Mohammad
  Babaeizadeh, Honghui Shi, Jianan Li, Shuicheng Yan, and Thomas~S Huang. 2016.
\newblock Seq-nms for video object detection.
\newblock \emph{arXiv preprint arXiv:1602.08465}.

\bibitem[{He et~al.(2016)He, Zhang, Ren, and Sun}]{he2016deep}
Kaiming He, Xiangyu Zhang, Shaoqing Ren, and Jian Sun. 2016.
\newblock Deep residual learning for image recognition.
\newblock In \emph{Proceedings of the IEEE conference on computer vision and
  pattern recognition}, pages 770--778.

\bibitem[{Hendricks et~al.(2017)Hendricks, Wang, Shechtman, Sivic, Darrell, and
  Russell}]{hendricks17iccv}
Lisa~Anne Hendricks, Oliver Wang, Eli Shechtman, Josef Sivic, Trevor Darrell,
  and Bryan Russell. 2017.
\newblock Localizing moments in video with natural language.
\newblock In \emph{Proceedings of the IEEE International Conference on Computer
  Vision (ICCV)}.

\bibitem[{Hendricks et~al.(2018)Hendricks, Wang, Shechtman, Sivic, Darrell, and
  Russell}]{hendricks2018localizing}
Lisa~Anne Hendricks, Oliver Wang, Eli Shechtman, Josef Sivic, Trevor Darrell,
  and Bryan Russell. 2018.
\newblock Localizing moments in video with temporal language.
\newblock In \emph{Proceedings of the 2018 Conference on Empirical Methods in
  Natural Language Processing}, pages 1380--1390.

\bibitem[{Hu et~al.(2017)Hu, Rohrbach, Andreas, Darrell, and
  Saenko}]{hu2017modeling}
Ronghang Hu, Marcus Rohrbach, Jacob Andreas, Trevor Darrell, and Kate Saenko.
  2017.
\newblock Modeling relationships in referential expressions with compositional
  modular networks.
\newblock In \emph{Computer Vision and Pattern Recognition (CVPR), 2017 IEEE
  Conference on}, pages 4418--4427. IEEE.

\bibitem[{Jhuang et~al.(2013)Jhuang, Gall, Zuffi, Schmid, and
  Black}]{jhuang2013towards}
Hueihan Jhuang, Juergen Gall, Silvia Zuffi, Cordelia Schmid, and Michael~J
  Black. 2013.
\newblock Towards understanding action recognition.
\newblock In \emph{Proceedings of the IEEE international conference on computer
  vision}, pages 3192--3199.

\bibitem[{Johnson et~al.(2015)Johnson, Krishna, Stark, Li, Shamma, Bernstein,
  and Fei-Fei}]{johnson2015image}
Justin Johnson, Ranjay Krishna, Michael Stark, Li-Jia Li, David Shamma, Michael
  Bernstein, and Li~Fei-Fei. 2015.
\newblock Image retrieval using scene graphs.
\newblock In \emph{Proceedings of the IEEE Conference on Computer Vision and
  Pattern Recognition}, pages 3668--3678.

\bibitem[{Karpathy et~al.(2014)Karpathy, Toderici, Shetty, Leung, Sukthankar,
  and Fei-Fei}]{KarpathyCVPR14}
Andrej Karpathy, George Toderici, Sanketh Shetty, Thomas Leung, Rahul
  Sukthankar, and Li~Fei-Fei. 2014.
\newblock Large-scale video classification with convolutional neural networks.
\newblock In \emph{CVPR}.

\bibitem[{Kazemzadeh et~al.(2014)Kazemzadeh, Ordonez, Matten, and
  Berg}]{kazemzadeh2014referitgame}
Sahar Kazemzadeh, Vicente Ordonez, Mark Matten, and Tamara~L Berg. 2014.
\newblock Referitgame: Referring to objects in photographs of natural scenes.
\newblock In \emph{EMNLP}, pages 787--798.

\bibitem[{Krishna et~al.(2017{\natexlab{a}})Krishna, Hata, Ren, Fei-Fei, and
  Niebles}]{krishna2017dense}
Ranjay Krishna, Kenji Hata, Frederic Ren, Li~Fei-Fei, and Juan~Carlos Niebles.
  2017{\natexlab{a}}.
\newblock Dense-captioning events in videos.
\newblock In \emph{International Conference on Computer Vision (ICCV)}.

\bibitem[{Krishna et~al.(2017{\natexlab{b}})Krishna, Zhu, Groth, Johnson, Hata,
  Kravitz, Chen, Kalantidis, Li, Shamma et~al.}]{krishna2017visual}
Ranjay Krishna, Yuke Zhu, Oliver Groth, Justin Johnson, Kenji Hata, Joshua
  Kravitz, Stephanie Chen, Yannis Kalantidis, Li-Jia Li, David~A Shamma, et~al.
  2017{\natexlab{b}}.
\newblock Visual genome: Connecting language and vision using crowdsourced
  dense image annotations.
\newblock \emph{International Journal of Computer Vision}, 123(1):32--73.

\bibitem[{Lei et~al.(2018)Lei, Yu, Bansal, and Berg}]{lei2018tvqa}
Jie Lei, Licheng Yu, Mohit Bansal, and Tamara Berg. 2018.
\newblock Tvqa: Localized, compositional video question answering.
\newblock In \emph{Proceedings of the 2018 Conference on Empirical Methods in
  Natural Language Processing}, pages 1369--1379.

\bibitem[{Li et~al.(2011)Li, Kulkarni, Berg, Berg, and Choi}]{li2011composing}
Siming Li, Girish Kulkarni, Tamara~L Berg, Alexander~C Berg, and Yejin Choi.
  2011.
\newblock Composing simple image descriptions using web-scale n-grams.
\newblock In \emph{Proceedings of the Fifteenth Conference on Computational
  Natural Language Learning}, pages 220--228. Association for Computational
  Linguistics.

\bibitem[{Li et~al.(2017)Li, Tao, Gavves, Snoek, Smeulders
  et~al.}]{li2017tracking}
Zhenyang Li, Ran Tao, Efstratios Gavves, Cees~GM Snoek, Arnold~WM Smeulders,
  et~al. 2017.
\newblock Tracking by natural language specification.
\newblock In \emph{CVPR}, volume~1, page~5.

\bibitem[{Lu et~al.(2014)Lu, Wu, and Chun~Zhu}]{lu2014online}
Yang Lu, Tianfu Wu, and Song Chun~Zhu. 2014.
\newblock Online object tracking, learning and parsing with and-or graphs.
\newblock In \emph{Proceedings of the IEEE Conference on Computer Vision and
  Pattern Recognition}, pages 3462--3469.

\bibitem[{Mao et~al.(2016)Mao, Huang, Toshev, Camburu, Yuille, and
  Murphy}]{mao2016generation}
Junhua Mao, Jonathan Huang, Alexander Toshev, Oana Camburu, Alan~L Yuille, and
  Kevin Murphy. 2016.
\newblock Generation and comprehension of unambiguous object descriptions.
\newblock In \emph{Proceedings of the IEEE conference on computer vision and
  pattern recognition}, pages 11--20.

\bibitem[{Mitchell et~al.(2013)Mitchell, Van~Deemter, and
  Reiter}]{mitchell2013generating}
Margaret Mitchell, Kees Van~Deemter, and Ehud Reiter. 2013.
\newblock Generating expressions that refer to visible objects.
\newblock In \emph{Proceedings of the 2013 Conference of the North American
  Chapter of the Association for Computational Linguistics}. Association for
  Computational Linguistics (ACL).

\bibitem[{Nagaraja et~al.(2016)Nagaraja, Morariu, and
  Davis}]{nagaraja2016modeling}
Varun~K Nagaraja, Vlad~I Morariu, and Larry~S Davis. 2016.
\newblock Modeling context between objects for referring expression
  understanding.
\newblock In \emph{European Conference on Computer Vision}, pages 792--807.
  Springer.

\bibitem[{Peng and Schmid(2016)}]{peng2016multi}
Xiaojiang Peng and Cordelia Schmid. 2016.
\newblock Multi-region two-stream r-cnn for action detection.
\newblock In \emph{European Conference on Computer Vision}, pages 744--759.
  Springer.

\bibitem[{Plummer et~al.(2017)Plummer, Wang, Cervantes, Caicedo, Hockenmaier,
  and Lazebnik}]{plummer2017flickr30k}
Bryan~A Plummer, Liwei Wang, Chris~M Cervantes, Juan~C Caicedo, Julia
  Hockenmaier, and Svetlana Lazebnik. 2017.
\newblock Flickr30k entities: Collecting region-to-phrase correspondences for
  richer image-to-sentence models.
\newblock \emph{International Journal of Computer Vision}, 123(1):74--93.

\bibitem[{Rodriguez et~al.(2008)Rodriguez, Ahmed, and
  Shah}]{rodriguez2008action}
Mikel~D Rodriguez, Javed Ahmed, and Mubarak Shah. 2008.
\newblock Action mach a spatio-temporal maximum average correlation height
  filter for action recognition.
\newblock In \emph{Computer Vision and Pattern Recognition, 2008. CVPR 2008.
  IEEE Conference on}, pages 1--8. IEEE.

\bibitem[{Rohrbach et~al.(2016)Rohrbach, Rohrbach, Hu, Darrell, and
  Schiele}]{rohrbach2016grounding}
Anna Rohrbach, Marcus Rohrbach, Ronghang Hu, Trevor Darrell, and Bernt Schiele.
  2016.
\newblock Grounding of textual phrases in images by reconstruction.
\newblock In \emph{European Conference on Computer Vision}, pages 817--834.
  Springer.

\bibitem[{Roy and Reiter(2005)}]{roy2005connecting}
Deb Roy and Ehud Reiter. 2005.
\newblock Connecting language to the world.
\newblock \emph{Artificial Intelligence}, 167(1-2):1--12.

\bibitem[{Russakovsky et~al.(2015)Russakovsky, Deng, Su, Krause, Satheesh, Ma,
  Huang, Karpathy, Khosla, Bernstein, Berg, and Fei-Fei}]{ILSVRC15}
Olga Russakovsky, Jia Deng, Hao Su, Jonathan Krause, Sanjeev Satheesh, Sean Ma,
  Zhiheng Huang, Andrej Karpathy, Aditya Khosla, Michael Bernstein,
  Alexander~C. Berg, and Li~Fei-Fei. 2015.
\newblock \href {https://doi.org/10.1007/s11263-015-0816-y} {{ImageNet Large
  Scale Visual Recognition Challenge}}.
\newblock \emph{International Journal of Computer Vision (IJCV)},
  115(3):211--252.

\bibitem[{Simonyan and Zisserman(2014)}]{simonyan2014two}
Karen Simonyan and Andrew Zisserman. 2014.
\newblock Two-stream convolutional networks for action recognition in videos.
\newblock In \emph{Advances in neural information processing systems}, pages
  568--576.

\bibitem[{Siskind(1990)}]{siskind1990acquiring}
Jeffrey~Mark Siskind. 1990.
\newblock Acquiring core meanings of words, represented as jackendoff-style
  conceptual structures, from correlated streams of linguistic and
  non-linguistic input.
\newblock In \emph{Proceedings of the 28th annual meeting on Association for
  Computational Linguistics}, pages 143--156. Association for Computational
  Linguistics.

\bibitem[{Soomro et~al.(2012)Soomro, Zamir, and Shah}]{soomro2012ucf101}
Khurram Soomro, Amir~Roshan Zamir, and Mubarak Shah. 2012.
\newblock Ucf101: A dataset of 101 human actions classes from videos in the
  wild.
\newblock \emph{arXiv preprint arXiv:1212.0402}.

\bibitem[{Tran et~al.(2015)Tran, Bourdev, Fergus, Torresani, and
  Paluri}]{tran2015learning}
Du~Tran, Lubomir Bourdev, Rob Fergus, Lorenzo Torresani, and Manohar Paluri.
  2015.
\newblock Learning spatiotemporal features with 3d convolutional networks.
\newblock In \emph{Proceedings of the IEEE international conference on computer
  vision}, pages 4489--4497.

\bibitem[{Venugopalan et~al.(2015)Venugopalan, Rohrbach, Donahue, Mooney,
  Darrell, and Saenko}]{venugopalan15iccv}
Subhashini Venugopalan, Marcus Rohrbach, Jeff Donahue, Raymond Mooney, Trevor
  Darrell, and Kate Saenko. 2015.
\newblock Sequence to sequence -- video to text.
\newblock In \emph{Proceedings of the IEEE International Conference on Computer
  Vision (ICCV)}.

\bibitem[{Vondrick et~al.(2013)Vondrick, Patterson, and
  Ramanan}]{vondrick2013efficiently}
Carl Vondrick, Donald Patterson, and Deva Ramanan. 2013.
\newblock Efficiently scaling up crowdsourced video annotation.
\newblock \emph{International Journal of Computer Vision}, 101(1):184--204.

\bibitem[{Wang et~al.(2016)Wang, Li, and Lazebnik}]{wang2016learning}
Liwei Wang, Yin Li, and Svetlana Lazebnik. 2016.
\newblock Learning deep structure-preserving image-text embeddings.
\newblock In \emph{Proceedings of the IEEE conference on computer vision and
  pattern recognition}, pages 5005--5013.

\bibitem[{Wolf et~al.(2014)Wolf, Lombardi, Mille, Celiktutan, Jiu, Dogan, Eren,
  Baccouche, Dellandr{\'e}a, Bichot et~al.}]{wolf2014evaluation}
Christian Wolf, Eric Lombardi, Julien Mille, Oya Celiktutan, Mingyuan Jiu, Emre
  Dogan, Gonen Eren, Moez Baccouche, Emmanuel Dellandr{\'e}a, Charles-Edmond
  Bichot, et~al. 2014.
\newblock Evaluation of video activity localizations integrating quality and
  quantity measurements.
\newblock \emph{Computer Vision and Image Understanding}, 127:14--30.

\bibitem[{Yamaguchi et~al.(2017)Yamaguchi, Saito, Ushiku, and
  Harada}]{yamaguchi2017spatio}
Masataka Yamaguchi, Kuniaki Saito, Yoshitaka Ushiku, and Tatsuya Harada. 2017.
\newblock Spatio-temporal person retrieval via natural language queries.
\newblock \emph{arXiv preprint arXiv:1704.07945}.

\bibitem[{Yang et~al.(2011)Yang, Teo, Daum{\'e}~III, and
  Aloimonos}]{yang2011corpus}
Yezhou Yang, Ching~Lik Teo, Hal Daum{\'e}~III, and Yiannis Aloimonos. 2011.
\newblock Corpus-guided sentence generation of natural images.
\newblock In \emph{Proceedings of the Conference on Empirical Methods in
  Natural Language Processing}, pages 444--454. Association for Computational
  Linguistics.

\bibitem[{Yin and Ordonez(2017)}]{yin2017obj2text}
Xuwang Yin and Vicente Ordonez. 2017.
\newblock Obj2text: Generating visually descriptive language from object
  layouts.
\newblock In \emph{Proceedings of the 2017 Conference on Empirical Methods in
  Natural Language Processing}, pages 177--187.

\bibitem[{Yu and Siskind(2013)}]{yu2013grounded}
Haonan Yu and Jeffrey~Mark Siskind. 2013.
\newblock Grounded language learning from video described with sentences.
\newblock In \emph{Proceedings of the 51st Annual Meeting of the Association
  for Computational Linguistics (Volume 1: Long Papers)}, volume~1, pages
  53--63.

\bibitem[{Yu et~al.(2018)Yu, Lin, Shen, Yang, Lu, Bansal, and
  Berg}]{yu2018mattnet}
Licheng Yu, Zhe Lin, Xiaohui Shen, Jimei Yang, Xin Lu, Mohit Bansal, and
  Tamara~L Berg. 2018.
\newblock Mattnet: Modular attention network for referring expression
  comprehension.
\newblock In \emph{Proceedings of the IEEE Conference on Computer Vision and
  Pattern Recognition (CVPR)}.

\bibitem[{Zhu et~al.(2017)Zhu, Wang, Dai, Yuan, and Wei}]{zhu2017flow}
Xizhou Zhu, Yujie Wang, Jifeng Dai, Lu~Yuan, and Yichen Wei. 2017.
\newblock Flow-guided feature aggregation for video object detection.
\newblock In \emph{Proceedings of the IEEE International Conference on Computer
  Vision}, volume~3.

\end{thebibliography}
\bibliographystyle{acl_natbib}

\end{document}